\documentclass[10pt,twocolumn,letterpaper]{article}

\usepackage{cvpr}
\usepackage{times}
\usepackage{epsfig}
\usepackage{graphicx}
\usepackage{amsmath}
\usepackage{amssymb}

\usepackage{booktabs}
\usepackage[breaklinks=true,bookmarks=false]{hyperref}

\cvprfinalcopy 

\def\cvprPaperID{****} 

\begin{document}

\title{Solving internal covariate shift in deep learning with linked neurons}

\author{Carles R. Riera Molina\\
    Universitat de Barcelona\\
    Gran Via de les Corts Catalanes 585\\
    08007 Barcelona, Spain\\
    {\tt\small crieramo8@alumnes.ub.edu}
    \and
    Oriol Pujol Vila\\
    Universitat de Barcelona\\
    Gran Via de les Corts Catalanes 585\\
    08007 Barcelona, Spain\\
    {\tt\small oriol\_pujol@ub.edu}
}

\maketitle

\begin{abstract}
    This work proposes a novel solution to the problem of internal covariate shift and dying neurons using the concept of linked neurons. We define the neuron linkage in terms of two constraints: first, all neuron activations in the linkage must have the same operating point. That is to say, all of them share input weights. Secondly, a set of neurons is linked if and only if there is at least one member of the linkage that has a non-zero gradient in regard to the input of the activation function. This means that for any input in the activation function, there is at least one member of the linkage that operates in a non-flat and non-zero area. This simple change has profound implications in the network learning dynamics. In this article we explore the consequences of this proposal and show that by using this kind of units, internal covariate shift is implicitly solved. As a result of this, the use of linked neurons allows to train arbitrarily large networks without any architectural or algorithmic trick, effectively removing the need of using re-normalization schemes such as Batch Normalization, which leads to halving the required training time. It also solves the problem of the need for standarized input data. Results show that the units using the linkage not only do effectively solve the aforementioned problems, but are also a competitive alternative with respect to state-of-the-art with very promising results.
\end{abstract}

\section{Introduction}

Training deep neural networks is a difficult task. It is not simply solving an optimization problem like support vector machines or gradient boosting. Factors such as the role of depth in the architecture, the nature of non-linear activation functions or the initialization of the weights are critical for the good behavior of the learning process. Over the years, the techniques, tricks, and hints used to train them have been refined, and nowadays there is a large variety of options aimed at improving the neural network learning process. 

The combination of depth and non-linear activations favors the change in the data distribution while it flows through the network. This effect is called internal covariate shift \cite{batchnorm} and lies at the core of problems such as vanishing/exploding gradients or dead neurons. Several solutions have appeared over the years. They either change the activation function to allow the gradient to flow back during the learning process (e.g. Leaky ReLU), or try to correct the internal distribution change by means of re-normalization techniques (e.g. Batch Normalization).

In this article, we propose the concept of {\it linked neurons}. Briefly, we define the neuron linkage in terms of two constraints: first, all neuron activations in the linkage must have the same operating point. In other words, all of them share input weights. Secondly, a set of neurons is linked if and only if there is at least one member of the linkage that has a non-zero gradient with respect to the input of the activation function. This means that for any input in the activation function, there is at least one member of the linkage that operates in a non-flat and non-zero area. In practice, this constraint forces at least a neuron in the linkage to operate in a complementary space to the rest of the set. This very simple concept has a dramatic change in network learning dynamics because it guarantees that even neurons with dead regions \footnote{We consider dead regions to be the part of the activation function that outputs a constant value. In this sense the gradient in those areas is zero and does not contribute to the learning process, inducing the dead neuron effect, or an effective reduction in the learning process.}, such as the rectified linear unit (ReLU), always receive gradient flow-back during the learning process. 

The concept of linked neurons has an important impact in the learning dynamics allowing to implicitly and effectively alleviate the problem of internal covariate shift, since the network will be able to learn disregarding any shift or data scaling in any layer. This has beneficial implications in the deep learning practice because it avoids the need of normalizing input data. Additionally, as internal covariate shift is implicitly handled it avoids the need of additional re-normalization techniques or robust activation functions. Seeing that this proposal ensures that there is always non-zero gradient flowing through the network, this opens new horizons for the deep learning practitioner, enabling the creation of new architectures with guaranteed learning dynamics.

Exploring similar ideas to the ones proposed in this article, we find CReLU\cite{crelu} and DReLU\cite{drelu}. These works focus on concatenating complementary ReLU activations, either in terms of mirroring or symmetry with respect to the origin, respectively. Differing from those works, which are focused on improving accuracy in convolutional neural networks or in recurrent neural networks, our proposal studies how the generalization of these works impacts on learning dynamics, effectively solving the problem of internal covariate shift and dead neurons regardless of the base activation function used. In particular, experiments in \cite{crelu} and \cite{drelu} use batch normalization. As we will show in our study, because both CReLU and DReLU can be seen as instantiations of the linked neurons concept, the use of Batch Normalization is redundant, because any linked neuron member implicitly corrects internal covariate shift. This can easily speed up the learning processing time by a factor of 2 to 4 without hindering the learning dynamics nor the predictive performance. 

In the following sections, we review the most widely used activation functions and methods to solve the problems described in the introduction. Following that, we formally introduce the concept of linked neurons. In the experimental section, we carry out an in-depth analysis of the proposal. First, the intuition behind the proposal and its dynamics is described and empirically shown on simple problems. Then, we study the learning behavior for different activations considering very wide and deep architectures. This will highlight the strengths and weaknesses of the techniques in different scenarios ranging from unnormalized data to lack of re-normalization in deep architectures. We will see that the proposed technique successfully handles all these scenarios. Finally, we compare the proposal in terms of performance metrics in different problems using state-of-the-art architectures, showing the suitability of the proposed technique.

\section{Background and state-of-the-art}\label{sec:review}

The problems of internal covariate shift and dying neurons both lie at the core of many problems in the learning process of deep learning models. In order to solve these effects two different approaches have been used in literature. One of those aims at alleviating these effects via changing the activation function. For example, dying neurons appear when, usually due to covariate shift, the distribution of data at the input of one layer sets the operation point of a neuron in a "flat" area of the activation function. This effect is clearly seen in activation functions such as ReLU, sigmoid, or hyperbolic tangent activations. In order to alleviate this effect, variations of these functions have been proposed. The proposals usually involve adding some slope in the activation function such that there is no flat area. Examples in this family of methods include Leaky ReLU\cite{leakyrelu} or Parametric ReLU\cite{prelu}. The second way for solving these problems is to directly deal with the covariate shift effect. Again, one may modify the activation function to try to deal with this effect. Scaled Exponential Rectified Linear Unit\cite{selu} (SELU) is the most successful example of this family. Alternatively, one can directly try to re-normalize the data output after each layer. The most well known method in this family is Batch Normalization\cite{batchnorm}. 

In this section we review all the previously named methods. We also include the newly proposed Swish activation \cite{swish} which is claimed to be superior to former activations.

\subsection{Rectified Linear Unit}

The Rectified Linear Unit is the first instantiation of a family of activation functions inspired in a simplification of the biological neurons activation \cite{relu_neurology}. It was firstly used in neural networks in \cite{firstrelu}. However, it is not until its apparition in \cite{alexnet}, where it is considered to be an integral part of its success, that it becomes popular. Its practical success has made it the default activation function for Computer Vision tasks. It consists in simply truncating the negative part of the input to zero, see Eq. \ref{eq:relu}.

\begin{equation}
f(x)={\begin{cases}x&{\mbox{if }}\quad x>0\\0&{\mbox{otherwise}}\end{cases}}
\label{eq:relu}
\end{equation}

It was originally formulated as a solution to the problem of vanishing gradient that commonly appeared when using sigmoid or hyperbolic tangent functions. ReLU effectively solves this problem by removing the saturation operation area when the input to the activation is a large positive value. This comes at the cost of allowing the appearance of exploding gradients. Because the positive side is not bounded, when the output is large, this value propagates back through the gradient, creating a feedback that in turn makes the activations even greater. This, ultimately ends in the exploding gradient effect and the the learning process breakdown. In order to solve this problem, practitioners frequently use the gradient clipping trick. Another equally important issue of ReLU is the death of the neuron effect. This occurs when the operation point of the activation function is deep on the negative side. This means that all activations turn zero independently of the input data. This effectively stops any possible recovery of the neuron during the learning process, rendering the neuron unusable. 

\subsection{Leaky Rectified Linear Units}

Leaky ReLU is chronologically, to the best of our knowledge, the first attempt at fixing the problem of dying neurons, \cite{leakyrelu}. As mentioned in the former subsection, since the cause of dying neurons is the truncation to zero of ReLU, Leaky ReLU replaces the truncation by a scaled linear model (the leak) see Eq. \ref{eq:leakyrelu}. This scaling allows the model to retain the non-linear behavior, simultaneously avoiding zero output activations. However, it introduces a hyperparameter to be tuned. This can be difficult to set because it is affected by the size of the activations, which in turn is determined by the amount of regularization, the learning rate and the distribution of the data, thus making it difficult to establish a safe default.

\begin{equation}
f(x)={\begin{cases}x&{\mbox{if }}x>0\\ax&{\mbox{otherwise}}\end{cases}}
\label{eq:leakyrelu}
\end{equation}

\subsection{Parametric ReLU}

A Leaky ReLU improvement is proposed under the name of Parametric ReLU, \cite{prelu}. It takes Leaky ReLU one step further by expanding the leak parameter, shared by all the neurons in the layer, into a full set of weights, one for each neuron. Contrary to Leaky ReLU, where the hyperparameter is fixed beforehand, in Parametric ReLU the weight is learnable and jointly trained with the network weights. Parametric ReLU has the following form, 

\begin{equation}
f(x_i)={\begin{cases}x_i&{\mbox{if }}\quad x_i>0\\ a_i x_i &{\mbox{otherwise}}\end{cases}}
\label{eq:parametricrelu}
\end{equation}
where the subscript $i$ in the activation identifies the elements of neuron $i$.

\subsection{Scaled Exponential Rectified Unit}

The SELU activation function is proposed in \cite{selu}. This activation attempts to simultaneously solve both the problem of dying neurons and covariate shift. To that end, the proposal introduces the use of an exponential function in the negative side of the activation, as in Exponential Linear Units \cite{ELU}, but improves it by adding a scaling term so that the activations are guaranteed to tend to a fixed point of mean 0 and variance 1. In practice, this can be seen as a pseudo-normalization of the data while if flows through the network. This proposal introduces a pair of parameters, $\alpha_{01} \approx 1.6733$ and $\lambda_{01} \approx 1.0507$, which are chosen to guarantee the attraction to the desired fixed point. As we will see in the experimental section, this is a very solid general purpose activation with the only caveat that it requires to normalize data beforehand to ensure that the operating point does not lie in the flat part of the activation function. The following equation gives the details of this activation function,

\begin{equation}
f(x)=\lambda{\begin{cases}
    x&{\mbox{if }}x>0\\
    \alpha \exp^x -\alpha&{\mbox{otherwise}}
    \end{cases}}
\label{eq:selu}
\end{equation}

\subsection{Swish}

Swish\cite{swish}  is a novel activation function which claims to be superior in terms of performance compared with the previous ones. It is equipped with an sigmoid multiplying the input in the negative side, which causes an small hump close to zero, see \ref{eq:swish}. In our understanding this small hump helps preventing dying neurons but its effect decreases as the input to the activation function is a large negative value. Additionally, it is worth noting that contrary to most activation functions\footnote{We do not consider in this set hyperbolic tangent and sigmoid which are also quasi-convex functions. However, differently from Swish, the quasi-convexity of these activations models a saturation effect.} Swish is not a convex function but a quasi-convex one. The effects of this change have not been specifically studied in the original article. The Swish activation function is as follows,

\begin{equation}
f(x)=\lambda{\begin{cases}
    x&{\mbox{if }}x>0\\
    x \cdot sigmoid(x) &{\mbox{otherwise}}
    \end{cases}}
\label{eq:swish}
\end{equation}

\subsection{Batch Normalization}

Batch Normalization \cite{batchnorm} is a technique developed with the aim of directly solving the problem of internal covariate shift. As its name suggests, it consists in normalizing each batch using a couple of normalization parameters corresponding to shift and scale for each unit. These parameters, namely $\gamma$ and $\beta$, are jointly trained with the weights using backpropagation as the rest of the network. This increases the number of parameters but improves the performance and enable higher learning rates.

\section{Linked neurons}\label{sec:proposal}

Our proposal aims to solve the problem of internal covariate shift, and thus avoid the dying neurons effect. Different from the solutions described in section \ref{sec:review}, our proposal can be considered as a framework as it will be valid for any of the former activation functions. This framework is formalized in terms of constraints that link a set of neurons. The idea behind the proposal is that in a set of linked neurons  there is always at least one neuron with an operating point outside a flat area. More formally, {\it we name a set of neurons to be linked if and only if there is at least one member of the linkage that has a non-zero gradient for any data point.} 

\begin{figure}[!h]
    \centering
    \begin{tabular}{c}
        \includegraphics[width=0.75\linewidth]{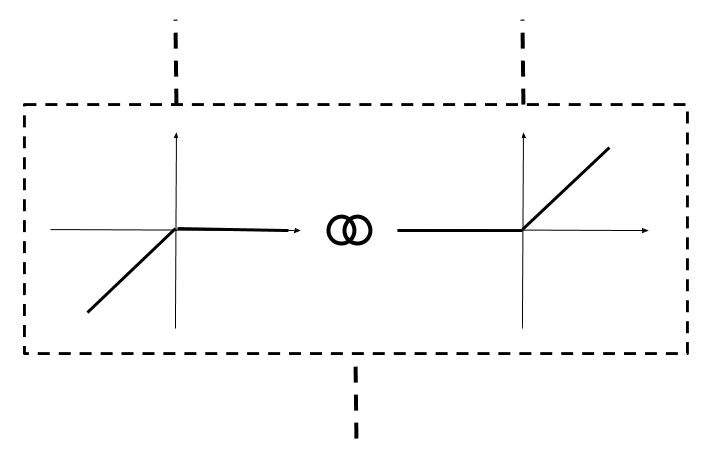}\\
        (a)\\
        \includegraphics[width=0.75\linewidth]{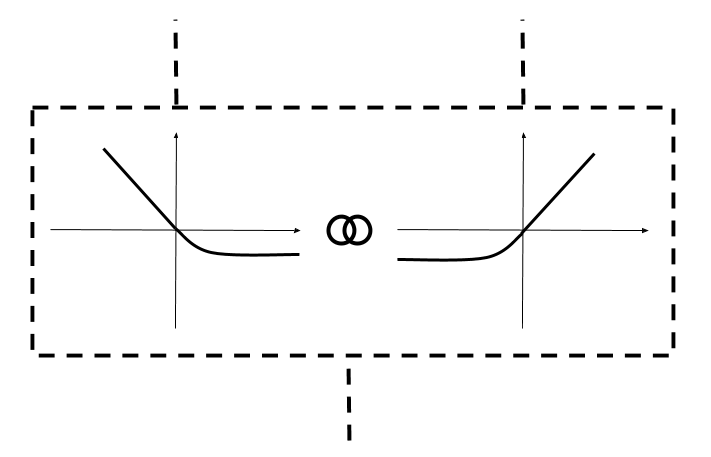}\\
        (b)\\
    \end{tabular}
    \caption{Two different examples of how to link two neurons: (a) shows two ReLUs coupled, (b) shows another option for coupling using SELU as base activation.}
    \label{fig:coupling}
\end{figure}

Although the amount of neurons linked can be an arbitrary number, for the sake of simplicity in the rest of the paper we will focus on the linking of just two neurons. This will allow us to use any of the already proposed activation units inside the proposed framework. Figure \ref{fig:coupling} shows two simple examples of linked neurons. Observe that all members of the coupling have exactly the same input. This is, all neurons in the coupling share the same weights. However, by construction, the activations must observe the constraint for the coupling, i.e. for any operating point (the value of the input to the activation) the coupling operates in a non-flat area (this guarantees a non-zero gradient). Figure \ref{fig:coupling}(a) shows two linked rectified linear units using a min-max policy for linking. Figure \ref{fig:coupling}(b) shows two linked SELU activations. Note that in this second case, we opted for using a different way of linking. In this case, horizontal mirroring fulfils the constraints. As expected, independently of the input value to the coupled set, there is always at least one of the activations functions active.  

Formally, $f\colon R^1\rightarrow R^{d}$, our linked activation, is a vectorial function with as many dimensions, $d$, as linked activations.  For the sake of our discussion, consider $g_i$ an element of a set of activations functions linked. Thus,  $f(z) = (g_1(z), \dots, g_d(z))$. Observe, that the input data to the vectorial function is the same for all $g_i$, with $z=w^Tx+w_0$, where $w$ are the input weights and $x$ is the data sample. Consider, now, the gradient of the loss function $\mathcal{L}$ with respect to the weights of that particular unit, i.e. $\nabla_w f(x)$. It is straightforward to proof that the gradient flowing back from the loss, $\mathcal{L}$ has the following form,

$$\nabla_w \mathcal{L}= \sum\limits_{i=1}^d  \frac{\partial \mathcal{L}}{\partial g_i(x)}\frac{\partial g_i(x)}{\partial w}$$

Observe that the gradient flowing back is a linear combination of the partial gradients of each of the linked units. If the linking condition holds, i.e. at least one of the partial gradient activations is different to zero, $\frac{\partial g_i(x)}{\partial w} \neq 0$, then $\nabla_w \mathcal{L}\neq0, \forall x, \frac{\partial \mathcal{L}}{\partial g_i(x)}\neq 0$.This directly ensures that there are no dead neurons during the training process and guarantees that there is a constant gradient flowing backwards coming from all units in the next layer.

There are many ways of defining activations such that they fulfill the linking conditions. If we restrict the linked activations to the case of just two neurons, for most current activation functions, horizontal mirroring (see Figure \ref{fig:coupling}(b)) suffices to fulfill the coupling constraints, this is
\begin{equation}
f(z) = (g(z),g(-z))
\label{eq:doublerelu}
\end{equation}
In the particular case of using the mirroring linkage with ReLU as base activation function, we recover the work of \cite{crelu}.

In the next section, we will specifically use this particular coupling in our experiments to show how the network dynamics change and check its behavior and performance.

\section{Experiments and results.}\label{sec:experiments}

We structure this section in two parts. In the first part, we investigate the effects of using linked neurons. We  study the effects of using linked neurons on wide and deep architectures separately. In the second part of the experiments, we evaluate the proposal in terms of performance accuracy using two state-of-the-art architectures, namely, AllCNN\cite{allcnn} and ResNet\cite{resnet}.

In all experiments we use the following single activations: ReLU, PReLU as a generalization of leaky ReLU, SELU, and Swish. We use the same activations using their linked versions using as a link the mirror linking as described in the proposal section. Because linked activations doubles the number of outputs, we effectively end up having the twice the number of parameters in the network. This has a clear implication in terms of an increment of capacity of the classifier and potentially in the increment of performance. For the sake of fairness in the experiments, we also compare the proposals with enlarged versions of the original network when appropriate. In those cases we propose to multiply the number of neurons of each layer by $\sqrt{2}$. This enlargement makes the experiments fair since the resulting network has exactly the same number of parameters as the the linked version. We have implemented all the experiments using Keras \cite{chollet2015keras} and Tensorflow \cite{tensorflow2015-whitepaper}. Code is available at \cite{code}

\subsection{Effect of linked neurons on wide and deep architectures.}

We study the effect of linked neurons in front of wide and deep architectures with proper initialization. We expect to see how the gap between our proposal and the competitors diminishes as the network gets wider. But, on the other hand, we would expect to see differences in deep networks where internal covariate shift can easily appear. 

\begin{itemize}
    \item For the width scenario, we train two one-layer networks of 50 and 400 convolutional units of size (3,3) each on the CIFAR-10 dataset. We use a batch size of 32 and train for 50 epochs (enough to ensure convergence). The results are shown in Table \ref{tab:onelayer}. 
    \item In order to explore the behavior of the different compared methods in terms of depth we use a very simple network. We designed a 50 layer network with 4 convolutional units per layer each of size (3,3) except for the first layer that has a size of (7,7). The network is trained using Adam \cite{adam} with a decay of 1e-6 and learning rate set to 1e-3. We have set the random seed in order to guarantee reproducibility. Results of this experiment is found in Table \ref{tab:basicnet}.
\end{itemize}

\begin{table*}[h]
    \centering
    
    \begin{tabular}{lrrrrrrrr}
	\toprule
	{} &    ReLU &   LK-ReLU &   PReLU &  LK-PReLU &    SELU &  LK-SELU &   Swish &  LK-Swish \\
	\midrule
	50 &  0.5767 &  0.5909 &  0.5771 &  0.5908 &  0.4992 &  0.5730 &  0.5559 &  0.5488 \\
	400 &  0.6154 &  0.6165 &  0.6156 &  0.6168 &  0.5227 &  0.599 &  0.5442 &  0.5405 \\
	\bottomrule
\end{tabular}
    
    \caption{Width experiment.}
    \label{tab:onelayer}
\end{table*}

Table \ref{tab:onelayer} shows the results of the width scenario. We observe that in general using a linked activation improves the results. This is expected as the number of parameters is increased with respect to the single activations. In terms of raw performance, Linked ReLU and Linked PReLU achieve the best performance. As it is also expected, when the number of neurons increases the difference between the single activation and the linked counterparts is reduced, most probably because both have enough discriminant capacity to solve the problem at the best of the architecture capabilities.

\begin{table}[h]
    \centering
    \begin{tabular}{lrr}
\toprule
{} &     Max &  Median \\
\midrule
ReLU             &  0.1000 &  0.1000 \\
LK-ReLU          &  0.5667 &  0.5607 \\
ReLU $\sqrt{2}$  &  0.1000 &  0.1000 \\
PReLU            &  0.1000 &  0.1000 \\
LK-PReLU         &  0.5458 &  0.5210 \\
PReLU $\sqrt{2}$ &  0.1000 &  0.1000 \\
SELU             &  0.5654 &  0.5643 \\
LK-SELU          &  0.5876 &  0.5670 \\
SELU $\sqrt{2}$  &  0.5707 &  0.5703 \\
Swish            &  0.1000 &  0.1000 \\
LK-Swish         &  0.1000 &  0.1000 \\
Swish $\sqrt{2}$ &  0.1000 &  0.1000 \\
BN               &  0.5202 &  0.5114 \\
\bottomrule
\end{tabular}

    \caption{Depth experiment.}
    \label{tab:basicnet}
\end{table}

Table  \ref{tab:basicnet} shows the results in the deep scenario. We compare basic activations with their linked counterparts. We also include extended versions of the basic activation in order to consider the extra number of parameters introduced by our proposal. As we can see, all linked versions of the activations except for Linked Swish are effectively able to learn despite the covariate shift. In particular Linked SELU achieves the best results, closely followed by Linked ReLU and Linked PReLU. We see how, as claimed in the original article, SELU is able to learn in this scenario. As we commented in the previous section, SELU is robust to internal covariate shift as long as it is operating in its correct range (normalization and weight initialization suffices to ensure that operating point). The failure of Linked Swish was not expected and deserves further study. Batch Normalization with ReLU is included for the sake of completion, but it cannot compete in terms of performance. The rest of the activations fail to learn. 

We also count the amount of dead neurons in this experiment. A dead neuron is characterized by having its activation equal to zero for all data samples. As expected, all Linked variants and SELU have all neurons alive. The only exception to this is Swish and Linked Swish that display an unexpected exploding gradient effect, effectively breaking their convergence despite lowering the learning rate. The rest of the activations fail to learn and show dead neurons from layer 7 onwards. This confirms the covariate shift negative effect in the learning process of deep architectures.

\subsection{Number of parameters}

\begin{table}[h]
    \centering
    \begin{tabular}{lrrr}
\toprule
{} &  Single &  $\sqrt{2}$ &  Linked \\
\midrule
ReLU  &  0.7337 &     0.7581 &  0.7696 \\
PReLU &  0.7201 &     0.7568 &  0.7687 \\
SELU  &  0.6981 &     0.7012 &  0.7559 \\
Swish &  0.6961 &     0.7376 &  0.7368 \\
\bottomrule
\end{tabular}

    \caption{Number of parameters accuracy comparison.}
    \label{tab:doubles}
\end{table}

Since our proposal has twice the number of outputs per neuron, this doubles the size of parameters needed at the output of a layer. Hereby, it is reasonable to suppose that this additional number of parameters increases the representation capabilities of the network, resulting in an increment of predictive performance. In this experiment we compare the performance when using each basic activation function with the linked counterpart and an extended version of the network which accounts for the extra parameters. The extended version of the network increases the number of units per layer $\sqrt{2}$ times the number of units, thus matching the number of parameters of the linked case. The network is a simple convolutional network composed by 4 layers of size (32, 32, 64, 64), plus a dense layer of 512 units. The dataset is CIFAR-10. We train the network for 200 epochs using RMSProp \cite{rmsprop} with a learning rate of $10^{-4}$ and a decay of $10^{-6}$. The results are shown in table \ref{tab:doubles}. Single stands for normal activations, linked for our proposal and  $\sqrt{2}$ for the extended versions. Notice how our proposal beats both single and  $\sqrt{2}$ counterparts in all the cases.

\subsection{Performance-focused experiments}

In this section we conduct experiments in order to maximize the predictive performance using state-of-the-art architectures. Their goal is to assess the viability of the linked neurons proposal and to strictly compare with the rest of the activations in the same scenario with the same experimental settings avoiding drifts from the stochastic nature of the optimization process. We use CIFAR-10 dataset and compare two very different architetures: AllCNN\cite{allcnn} and ResNet\cite{resnet}.

{\bf Architecture details and experimental settings:}
\begin{table*}[h]
    \centering
    
    \begin{tabular}{rrrrrrrr}
\toprule
ReLU &  LK-ReLU &   PReLU &  LK-PReLU &    SELU &  LK-SELU &   Swish &  LK-Swish \\
\midrule
0.9301 &   0.9468 &  0.9422 &    0.9232 &  0.9304 &   0.9197 &  0.9261 &    0.8971 \\
\bottomrule
\end{tabular}

    \caption{Allcnn experiment.}
    \label{tab:allcnn}
\end{table*}

\begin{table*}[h]
    \centering
    
    \begin{tabular}{rrrrrrrr}
\toprule
   ReLU &   LReLU &   PReLU &  LPReLU &    SELU &   LSELU &  Swish &  LSwish \\
\midrule
 0.8973 &  0.9027 &  0.8907 &  0.9027 &  0.9041 &  0.9134 &  0.905 &  0.9139 \\
\bottomrule
\end{tabular}

    \caption{ResNet50 experiment.}
    \label{tab:resnet50}
\end{table*}

\begin{table*}[!ht]
    \centering
    
    \begin{tabular}{lrrrrrr}
\toprule
{} &  LK-ReLU BN &  LK-ReLU No BN &   SELU BN &  SELU No BN &  LK-SELU BN &  LK-SELU No BN \\
\midrule
Accuracy & 0.9027 &0.9048 &0.9041 &0.8611 & 0.9134 &0.9088 \\
Time ratio &    2.09 &         1 &  2.07 &    1.38 &    2.08 &       1.6 \\
\bottomrule
\end{tabular}

    \caption{ResNet50 without Batch Normlization experiment.}
    \label{tab:resnet50_nobatchnorm}
\end{table*}

\begin{itemize}
    \item AllCNN \cite{allcnn} is a network architecture that currently holds performance figures very close to the best performant state-of-the-art methods in CIFAR-10. The network is rather wide featuring 10 layers of up to 196 units each. We followed the recommended settings. However, because the linked versions usually accommodate larger gradient magnitude values flowing back we changed the learning rate policy and use the following step approach: we keep the initial learning rate for 500 epochs, then $0.75 lr$ until epoch 2000, $ 0.5 lr$ until epoch 2350, $0.1 lr$ until epoch 2750, and, finally,  $0.05 lr$ until epoch 3000. The original values for the learning rates are $lr \in \{0.01, 0.001\} $. We also report results using a fixed learning rate $lr = 0.01$. We use SGD with Nesterov momentum for optimization. We apply a mild data augmentation consisting of random flips, and vertically and horizontal shifts of $10\%$ of the corresponding size axis.
    
    \item
    ResNet is the family of networks based on residual connections \cite{resnet}. Residual connections allow to significantly increase the depth of the network while also increasing the performance. In this experiment we use a variation from ResNet50, originally designed for ImageNet, scaled down in order to suit CIFAR-10. The architecture features 50 convolutional layers of up to 128 units. The network is trained it using Adam \cite{adam} with a learning rate of 0.001 and a decay of 1e-6 for 2000 epochs. As in the former case, we apply a mild data augmentation consisting of random flips, and vertically and horizontal shifts of $10\%$ of the corresponding size axis. 
\end{itemize}

The results are presented in tables \ref{tab:allcnn} and \ref{tab:resnet50}. It is worth noting that in terms of raw performance, linked neuron proposals achieve the best results. In the case of AllCNN, Linked ReLU achieves the best accuracy, closely followed by PReLU. However, we observe a lack of consistency in the results of the linked versions. We believe that this is due to the different gradient dynamics intrinsic to the linked versions. The specific study of this effect deserves further attention. In ResNet50, Linked SeLU outperforms the rest of the activations. However, in this case all linked versions achieve better results that their basic counterparts. 

Considering the point that both our proposal and SELU share the property of being robust to covariate shift, we propose another experiment in which we use the same ResNet network but removing its Batch Normalization layers. We consider LK-ReLU, LK-SELU and SELU.  Results are presented in Table \ref{tab:resnet50_nobatchnorm}

As we can see, LK-ReLU and LK-SELU perform similarly with or without Batch Normalization. Both SELU and LK-SELU without Batch Normalization are very sensitive to the learning rate. This seems to suggest that this might be related to the intrinsic dynamics of SELU. We additionally compare the convergence speed. Taking as reference LK-RELU without Batch Normalization, we can see in Table \ref{tab:resnet50_nobatchnorm} that Linked versions are up to 2 times faster than using Batch Normalization. If we compare the speed gain achieved when dropping Batch Normalization we find that LK-RELU is 2.08 times faster, SELU is 1.50 times faster, and LK-SELU 1.3 times faster. 

\section{Conclusion, advantages and future work}\label{sec:conclusions}

In this article we have introduced the concept of {\it linked neurons}. Linked neurons are a set of neurons joined in terms of two constraints. The first constraint states that all elements in the linkage share the same input. In practical terms, this is accomplished by sharing weights. The second constraint forces that at least one of the activation functions in the linkage operates in a locally non-constant region of the function. This ensures that any non-zero gradient that will propagate backwards will not vanish. As a result, network learning dynamics change, ensuring that there are no dead neurons and providing a constant gradient flow in the learning process that implicitly corrects any internal covariate shift. This can help the deep learning practitioner ensuring that the model will learn even in front of not properly normalized data. Additionally, learning dynamics adapts the network to achieve internal covariate shift invariance. This opens the possibility to remove re-normalization schemes such as Batch Normalization from the network, considerably speeding up the learning process. We believe that thanks to the regularity in the gradient flow of our approach this opens the opportunity to explore new learning rate policies that guarantee avoiding exploding gradients and allowing to effectively increase the convergence rate of the algorithm. We also think that this approach may help in the definition of new architectures focused on the concept of linkages among neurons. As a side note, most of the tips and tricks, and good practices in neural networks are specifically tuned to independent activations. These must be updated to the linked neurons scenario.

{\small
\bibliographystyle{ieee}
\bibliography{egbib}
}

\end{document}